\documentclass{article}
\usepackage{arxiv}
\usepackage[utf8]{inputenc} 
\usepackage[T1]{fontenc}    
\usepackage{hyperref}       
\usepackage{url}            
\usepackage{booktabs}       
\usepackage{amsfonts}       
\usepackage{nicefrac}       
\usepackage{microtype}      
\usepackage{lipsum}		    
\usepackage{graphicx}
\usepackage{doi}
\usepackage{glossaries}
\usepackage[table]{xcolor}
\definecolor{green}{rgb}{0,.5,0}
\definecolor{magenta}{rgb}{.75,0,.75}
\usepackage{soul}
\sethlcolor{lightgray}
\usepackage[labelfont=bf,font=footnotesize]{caption}
\usepackage{float}
\usepackage{multicol}
\usepackage{wrapfig} 
\usepackage{csquotes} 
\usepackage{tikz}
\newcommand*{\rootpath}{}

\newcommand\scale{1} 
\newcommand\scaleppt{0.65}
\newcommand\scalepy{0.35}
\title{Convolutional neural networks for medical image segmentation}
\renewcommand{\undertitle}{A closer look at the spatial origin of the data and highlighting the importance of sampling}
\renewcommand{\headeright}{}
\renewcommand{\shorttitle}{Convolutional neural networks for medical image segmentation}
\author{
    \href{https://orcid.org/0000-0001-7206-2671}{\includegraphics[scale=0.06]{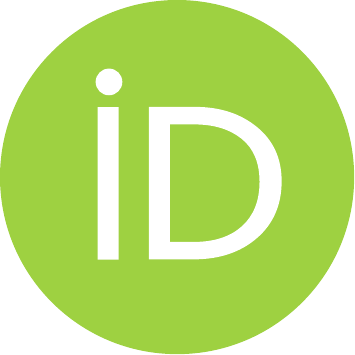}\hspace{1mm}Jeroen Bertels}\\
	Processing Speech and Images\\
	Department of Electrical Engineering\\
	KU Leuven, Belgium\\
	\texttt{jeroen.bertels@kuleuven.be}\\
	\And
	David Robben\\
	Processing Speech and Images\\
	Department of Electrical Engineering\\
	KU Leuven, Belgium\\
	\texttt{david.robben@kuleuven.be}\\
    \And
	Robin Lemmens\\
	Laboratory of Neurobiology\\
	Department of Neurosciences\\
	KU Leuven, Belgium\\
	\texttt{robin.lemmens@kuleuven.be}
    \And
	Dirk Vandermeulen\\
	Processing Speech and Images\\
	Department of Electrical Engineering\\
	KU Leuven, Belgium\\
	\texttt{dirk.vandermeulen@kuleuven.be}\\
}
\begin{document}
\date{}  
\maketitle
\newglossaryentry{cbf}{name={CBF},description={cerebral blood flow}}
\newglossaryentry{cbv}{name={CBV},description={cerebral blood volume}}
\newglossaryentry{cpp}{name={CPP},description={cerebral perfusion pressure}}
\newglossaryentry{cvr}{name={CVR},description={cerebrovascular resistance}}
\newglossaryentry{oef}{name={OEF},description={oxygen extraction fraction}}
\newglossaryentry{ais}{name={AIS},description={acute ischemic stroke}}
\newglossaryentry{tso}{name={TsO},description={time since onset}}
\newglossaryentry{ttt}{name={TtT},description={time to treatment}}
\newglossaryentry{tici}{name={mTICI},description={modified Thrombolysis in Cerebral Infarction}}
\newglossaryentry{ivt}{name={IVT},description={intravenous treatment}}
\newglossaryentry{iat}{name={IAT},description={intra-arterial treatment}}
\newglossaryentry{evt}{name={EVT},description={endovascular treatment}}
\newglossaryentry{mrs}{name={mRS},description={modified Rankin Scale}}
\newglossaryentry{nihss}{name={NIHSS},description={National Institute of Health Stroke Scale}}
\newglossaryentry{ich}{name={ICH},description={intracranial hemorrhage}}
\newglossaryentry{ct}{name={CT},description={computed tomography}}
\newglossaryentry{ncct}{name={NCCT},description={non-enhanced or non-contrast CT}}
\newglossaryentry{aspects}{name={ASPECTS},description={Alberta Stroke Program Early CT Score}}
\newglossaryentry{mca}{name={MCA},description={middle cerebral artery}}
\newglossaryentry{mri}{name={MRI},description={magnetic resonance imaging}}
\newglossaryentry{dwi}{name={DWI},description={diffusion weighted imaging}}
\newglossaryentry{adc}{name={ADC},description={apparent diffusion coefficient}}
\newglossaryentry{ctp}{name={CTP},description={CT perfusion}}
\newglossaryentry{cta}{name={CTA},description={CT angiography}}
\newglossaryentry{pwi}{name={PWI},description={perfusion-weighted imaging}}
\newglossaryentry{tcc}{name={TCC},description={time concentration curve}}
\newglossaryentry{irf}{name={irf},description={impulse response function}}
\newglossaryentry{imp}{name={imp},description={Dirac impulse}}
\newglossaryentry{aif}{name={AIF},description={arterial imput function}}
\newglossaryentry{pet}{name={PET},description={positron emission tomography}}
\newglossaryentry{spect}{name={SPECT},description={single photon emission CT}}
\newglossaryentry{mra}{name={MRA},description={MRI angiography}}
\newglossaryentry{cc}{name={CC},description={collateral circulation}}
\newglossaryentry{3d}{name={3D},description={3 dimensional}}
\newglossaryentry{alara}{name={ALARA},description={as low as reasonably possible}}
\newglossaryentry{dcv}{name={DCV},description={deconvolution}}
\newglossaryentry{cnn}{name={CNN},description={convolutional neural network}}
\newglossaryentry{tmax}{name={Tmax},description={location of maximum of TCC\textsubscript{irf}}}
\newglossaryentry{rcbf}{name={rCBF},description={relative CBF}}
\newglossaryentry{cbct}{name={CBCT},description={C-arm or cone-beam CT}}
\newglossaryentry{cbctp}{name={CBCTP},description={CBCT perfusion}}
\newglossaryentry{de}{name={DE},description={dual-energy}}
\newglossaryentry{dencct}{name={DENCCT},description={DE NCCT}}
\newglossaryentry{rcbv}{name={rCBV},description={relative CBV}}
\newglossaryentry{o24h}{name={O24h},description={occlusion present at follow-up}}
\newglossaryentry{flair}{name={FLAIR},description={fluid-attenuated inversion recovery MRI}}
\newglossaryentry{vof}{name={VOF},description={venous output function}}
\newglossaryentry{glm}{name={GLM},description={generalized linear model}}
\newglossaryentry{csf}{name={CSF},description={cerebrospinal fluid}}
\newglossaryentry{minip}{name={minIP},description={minimum intensity projection}}
\newglossaryentry{meanip}{name={meanIP},description={mean intensity projection}}
\newglossaryentry{maxip}{name={maxIP},description={maximum intensity projection}}
\newglossaryentry{relu}{name={ReLU},description={rectified linear unit}}
\newglossaryentry{ica}{name={ICA},description={internal carotid artery}}
\newglossaryentry{aca}{name={ACA},description={arterior cerebral artery}}
\newglossaryentry{rep+}{name={Rep$^+$},description={(the group of) reperfusers}}
\newglossaryentry{rep-}{name={Rep$^-$},description={(the group of) non-reperfusers}}
\newglossaryentry{rep+-}{name={Rep$^{+/-}$},description={combination of Rep$^+$ and Rep$^-$}}
\newglossaryentry{lps}{name={LPS},description={left posterior superior}}
\newglossaryentry{mse}{name={MSE},description={mean squared error}}
\newglossaryentry{adv}{name={$|\Delta\mathrm{V}|$},description={absolute volume error}}
\newglossaryentry{rf}{name={RF},description={receptive field}}
\newglossaryentry{prf}{name={pRF},description={physical receptive field}}
\newglossaryentry{ce}{name={CE},description={cross-entropy}}
\newglossaryentry{sgd}{name={SGD},description={stochastic gradient descent}}
\newglossaryentry{dsc}{name={DSC},description={Dice coefficient}}
\newglossaryentry{dv}{name={$\Delta\mathrm{V}$},description={volume bias or error}}
\newglossaryentry{hd95}{name={HD95},description={95th percentile Hausdorff distance}}
\newglossaryentry{ppv}{name={PPV},description={precision or positive predictive value}}
\newglossaryentry{tpr}{name={TPR},description={recall or true positive rate}}
\newglossaryentry{ece}{name={ECE},description={expected calibration error}}
\newglossaryentry{auc}{name={AUC},description={area under the precision-recall curve}}
\newglossaryentry{fov}{name={FOV},description={field of view}}
\newglossaryentry{m}{name={M},description={MRCLEAN}}
\newglossaryentry{c}{name={C},description={CRISP}}
\newglossaryentry{k}{name={K},description={KAROLINSKA}}
\newglossaryentry{mck}{name={MCK},description={combination of M, C and K}}
\newglossaryentry{clpr}{name={ClPr},description={clinical practice}}
\newglossaryentry{jl}{name={JL},description={Julie Lambert}}
\newglossaryentry{jd}{name={JD},description={Jelle Demeestere}}
\newglossaryentry{grt}{name={GrT},description={grow time}}
\newglossaryentry{bao}{name={BAO},description={basilar artery occlusion}}
\newglossaryentry{rtpa}{name={rtPA},description={recombinant tissue plasminogen activator}}
\newglossaryentry{nn}{name={NN},description={neural network}}
\newglossaryentry{gpu}{name={GPU},description={graphics processing unit}}
\newglossaryentry{rgb}{name={RGB},description={red green blue}}
\newglossaryentry{pca}{name={PCA},description={principle component analysis}}
\newglossaryentry{fcn}{name={FCN},description={fully-convolutional network}}
\newglossaryentry{dl}{name={DL},description={Dice loss}}
\newglossaryentry{sd}{name={SD},description={soft Dice}}
\newglossaryentry{gan}{name={GAN},description={generative adversarial network}}
\newglossaryentry{t1w}{name={T1w},description={T1-weighted MRI}}
\newglossaryentry{stn}{name={STN},description={spatial transformer network}}
\newglossaryentry{dvn2}{name={DVN2},description={DeepVoxNet2}}
\newglossaryentry{gt}{name={GT},description={ground truth}}
\newglossaryentry{vs}{name={VS},description={voxel size}}
\newglossaryentry{lvo}{name={LVO},description={large vessel occlusion}}
\newglossaryentry{nexis}{name={NEXIS},description={NExt generation X-ray Imaging System}}
\newglossaryentry{ip}{name={IP},description={intensity projection}}
\begin{abstract}
In this article, we look into some essential aspects of convolutional neural networks (\gls{cnn}s) with the focus on medical image segmentation. First, we discuss the \gls{cnn} architecture, thereby highlighting the spatial origin of the data, voxel-wise classification and the receptive field. Second, we discuss the sampling of input-output pairs, thereby highlighting the interaction between voxel-wise classification, patch size and the receptive field. Finally, we give a historical overview of crucial changes to \gls{cnn} architectures for classification and segmentation, giving insights in the relation between three pivotal \gls{cnn} architectures: FCN~\cite{Long2015}, U-Net~\cite{Ronneberger2015} and DeepMedic~\cite{Kamnitsas2017}.
\end{abstract}
\section{The \gls{cnn} basics}
\begin{figure}[b]
    \newcommand\myscale{\scaleppt} 
    \setlength\tabcolsep{0pt}
    \centering
    \includegraphics[scale=\scale,scale=\myscale]{\rootpath 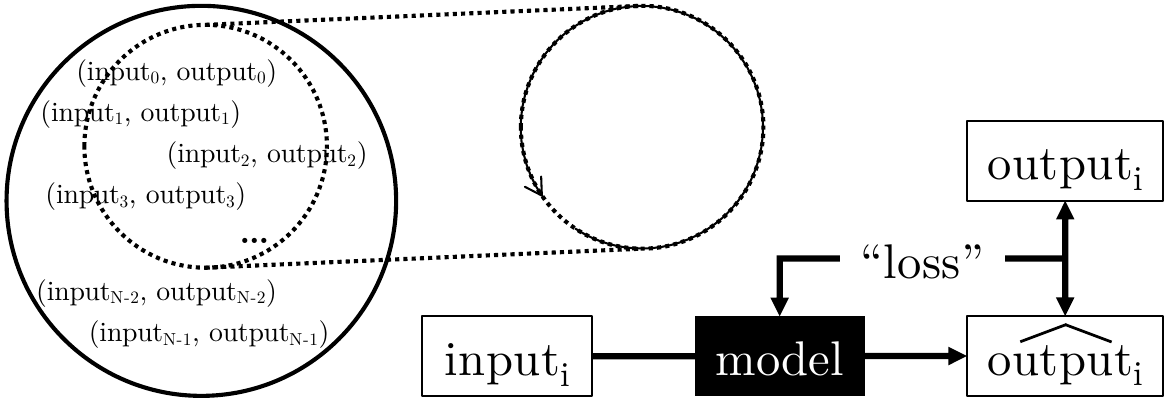}
    \caption[A schematic representation of data-driven machine learning.]{A schematic representation of machine learning in the context of data-driven model optimization. The model, e.g., the \gls{cnn}, has trainable parameters that are tuned iteratively to minimize the loss between the observed and predicted output. To do so, a sampling scheme is pivotal to select subsets out of a dataset of input-output pairs.}
    \label{fig:machine_learning}
\end{figure}
\begin{figure}[t]
    \newcommand\myscale{\scaleppt} 
    \setlength\tabcolsep{0pt}
    \centering
    \begin{tabular}{cc}
        \includegraphics[scale=\scale,scale=\myscale,trim={0 0 -1cm 0},clip]{\rootpath 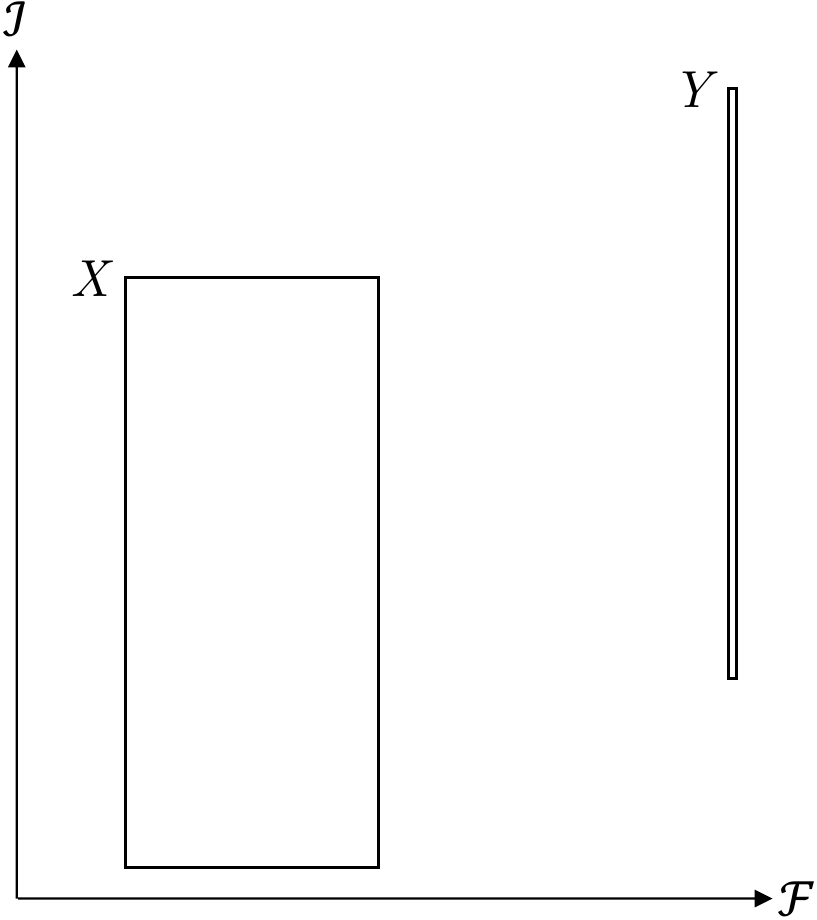}&
        \includegraphics[scale=\scale,scale=\myscale]{\rootpath 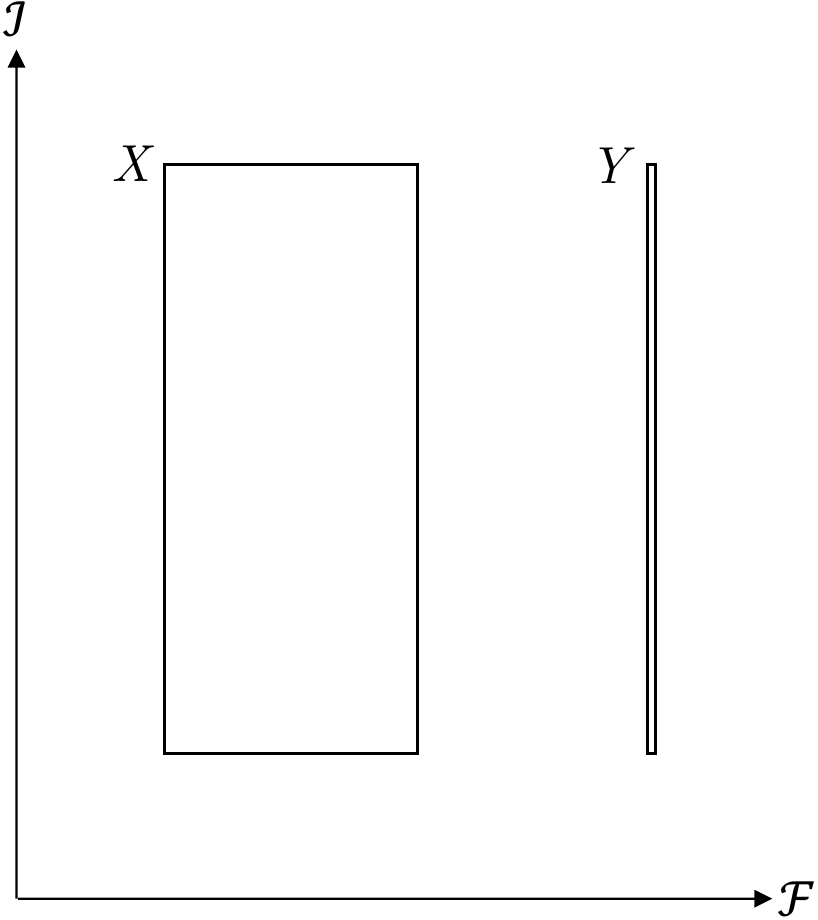}
    \end{tabular}
    \caption[Schematic representation of $X$ and $Y$.]{Schematic representation of $X$ and $Y$ as being so-called feature maps. While they are not necessarily aligned (LEFT), in a typical CNN-based segmentation setting they are (RIGHT).}
    \label{fig:abstraction}
\end{figure}

Often we can view the training of \gls{cnn}s as a model that learns a mapping from an input to an output that minimizes the loss over a given set of examples (Figure~\ref{fig:machine_learning}). In the context of medical image segmentation, one could for example define the input $X$ to be a 3D computed tomography (\gls{ct}) image and the output $Y$ to be the manual segmentation of say a tumor. This way, we can view both $X$ and $Y$ as so-called \textit{feature maps}, i.e., multi-dimensional arrays that have a number of spatial dimensions and a feature dimension. Grouping the spatial dimensions all under a, somewhat abstract, axis $\mathcal{I}$ and taking as feature axis $\mathcal{F}$, this is depicted in Figure~\ref{fig:abstraction}. In the previous example the images are aligned, but when $Y$ represents another medical image, often image registration is applied first to align the two. As a result, the spatial indices $i \in \mathcal{I}$ present in $Y$ are also present in $X$. The task of the \gls{cnn} in such a segmentation setting then boils down to mapping features onto one another, thereby keeping spatial correspondence. In this case, a spatial index refers to a pixel (2D) or voxel (3D) and the feature represents \textit{the information} (e.g., X-ray attenuation) captured by the imaging modality when interacting with that \textit{piece of tissue} (e.g., a voxel of 1~mm~x~1~mm~x~1~mm brain tissue contains thousands of cells and hundreds capillaries) at that spatial location. 
\subsection{Successive feature mapping}
Now take $x$ and $y$ being the \gls{ct} and manual tumor segmentation of a particular subject. We can view the \gls{cnn} as a mapping function $m$, parametrized by a set of weights (also called parameters) $\Omega$, that needs to transform $x$ into $\hat{y}$, i.e., the prediction of $y$:
\begin{equation}
    m(x)=\hat{y}.
    \label{eq:m}
\end{equation}
While it is important to make a distinction between $y$ and $\hat{y}$, and usually $\hat{y} \ne y$, for now (to simplify notation) we can assume $m$ to perform the perfect mapping function, i.e., $\hat{y}=y$. Typical for a \gls{cnn}, is that we can view $m$ as the successive application of L layer functions $m^l$, each parametrized by its own set of weights $\Omega^l$: 
\begin{equation}
    m=m^\mathrm{L}\circ m^{\mathrm{L}-1}\circ ...\circ m^2\circ m^1. 
    \label{eq:mls_}
\end{equation}
In essence, every $m^l$ performs a mapping $m^l(x^l)=y^l$, with $x^l$ and $y^l$ defined as the input and output of layer $l$, respectively. As such, it is trivial to see that $x=x^1$, $y^{l-1}=x^l$ and $y^\mathrm{L}=y$. Furthermore, it is clear that every combination of successive layers can be seen as a \gls{cnn}, with in the limit every $m^l$ being a \gls{cnn} on its own.
\subsection{Pattern recognition}
To understand how many successive mapping functions, or layers, are needed in a practical \gls{cnn}, let us start with having a look at the internal mechanisms of any mapping function $m^l$. What if we would view any feature map $y^l$ as being a spatial distribution of features, representing how much a certain pattern $f$ was present at a particular spatial index $i$ (a voxel if $\mathcal{I}$ is \gls{3d}) in $x^l$? Indeed, this is also true for our input $x$, the \gls{ct}, and our output $y$, the manual tumor segmentation. A simple way to use the internal weights $\Omega^l$ of any $m^l$ to detect a certain pattern, would be to have $\Omega^l$ containing the pattern itself and taking the dot product at any index $i$ to compute the linear activation $a^l$ using:
\begin{equation}
    a^l_i = \sum_\phi^{\Phi^l(i)} \mathrm{W}_\phi^l x^l_\phi + \mathrm{w}^l.
    \label{eq:mlmc}
\end{equation}
We notice four important aspects. First, performing the computation at every spatial index performs a cross-correlation of the values in $x^l$ and $\Omega^l=(\mathrm{W}^l,\mathrm{w}^l)$ or a convolution if one of them is considered flipped, hence the name \gls{cnn}. Second, the entire feature dimension of the previous layer output $y^{l-1}=x^l$ is reduced and thus patterns are looked for and allowed to be distributed across the entire feature axis. Third, patterns are looked for locally in a spatial neighborhood $\Phi^l$ around $i$. Fourth, the number of features in the new activation map $a^l$ depends on the shapes of $\mathrm{W}^l$ and $\mathrm{w}^l$. A final ingredient of $m^l$ is that it uses a so-called non-linear activation function $\sigma^l$ to avoid the entire mapping represented by $m$ to be linear:
\begin{equation}
    y^l_i = \sigma^l(a_i^l).
    \label{eq:mlmc2}
\end{equation}
The non-linear activation function is not per se element-wise (e.g., the softmax function), but it keeps the shape of $a^l$ for $y^l$. We further want to note that $\sigma$ is typically monotonously increasing, hence the idea that, the higher the feature value, the more the pattern was present, still holds, and that this entire understanding to be true requires appropriate normalization. 
\subsection{Voxel-wise classification}\label{sec:voxel-wise_classification}
\begin{figure}[b]
    \newcommand\myscale{\scaleppt} 
    \setlength\tabcolsep{0pt}
    \centering
    \includegraphics[scale=\scale,scale=\myscale]{\rootpath 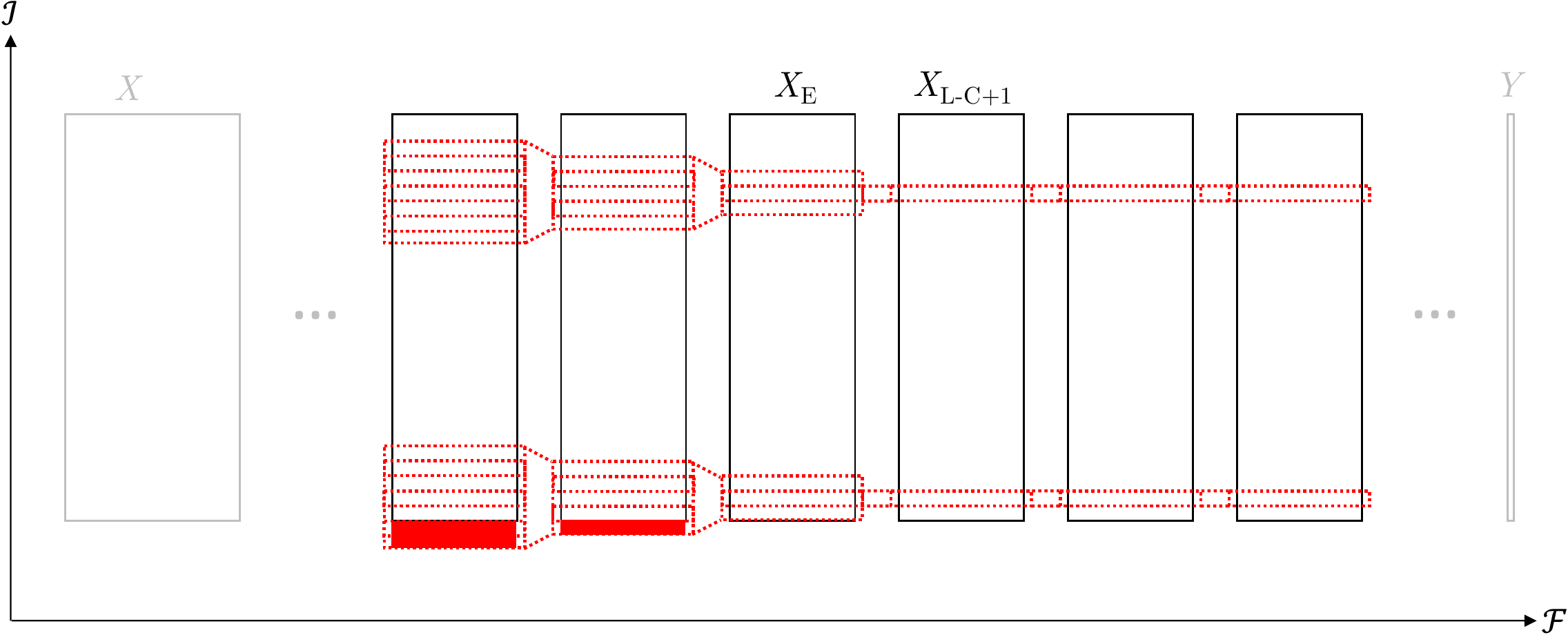}
    \caption[Voxel-wise classification, extraction and the receptive field.]{A schematic representation of voxel-wise classification, feature extraction and the receptive field. The first layers are typically viewed as feature extraction layers that look for patterns inside the receptive field of the \gls{cnn} in $X$. The final layers can be seen as performing voxel-wise classification.}
    \label{fig:receptive_field}
\end{figure}
\begin{figure}[t]
    \newcommand\myscale{\scalepy} 
    \setlength\tabcolsep{0pt}
    \centering
    \begin{tabular}{c|c}
        Effect of width $\downarrow$ & Effect of depth $\downarrow$\\
        \includegraphics[scale=\scale,scale=\myscale,trim={0 -1cm -0.5cm 6cm},clip]{\rootpath 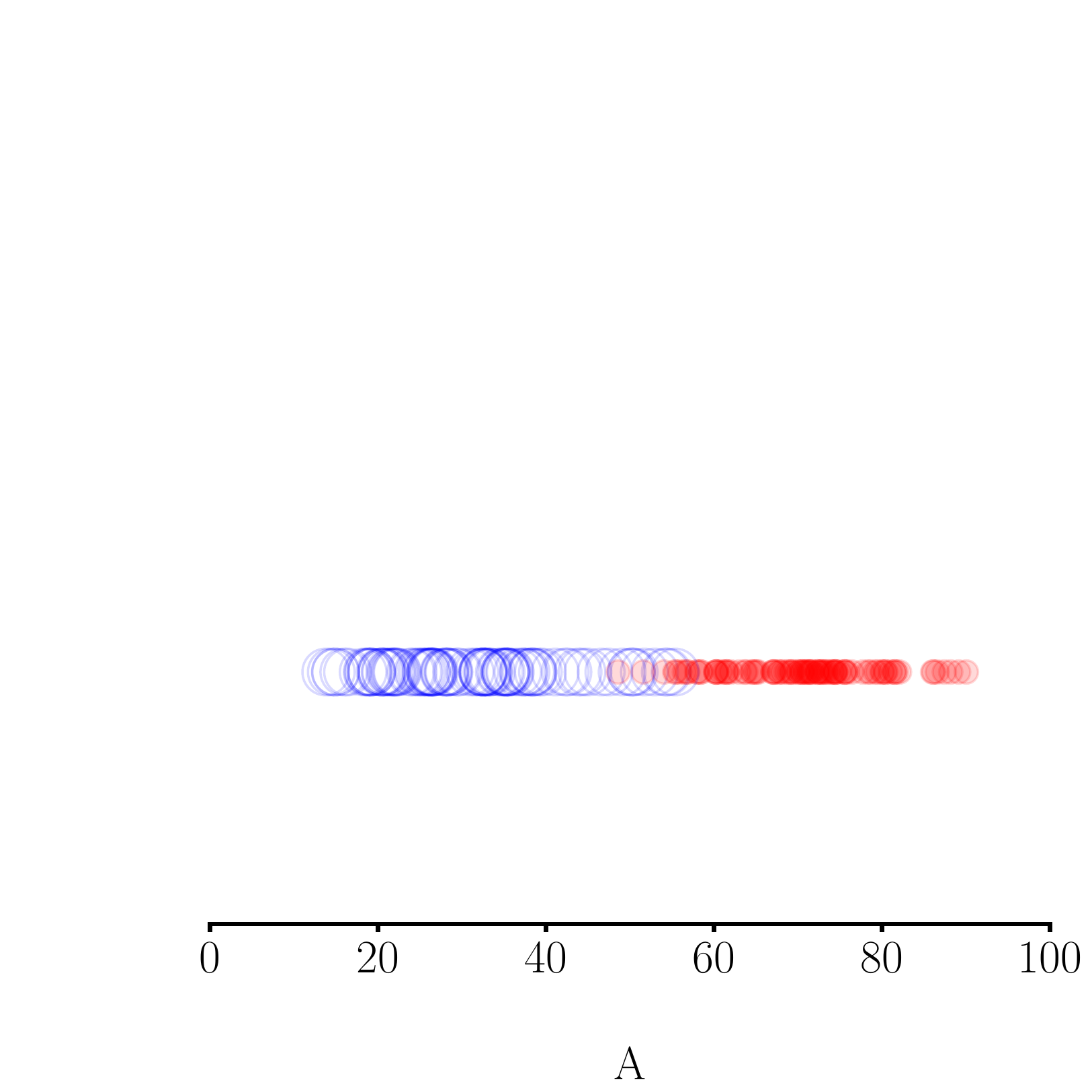}&
        \includegraphics[scale=\scale,scale=\myscale,trim={-0.5cm -1cm 0 6cm},clip]{\rootpath 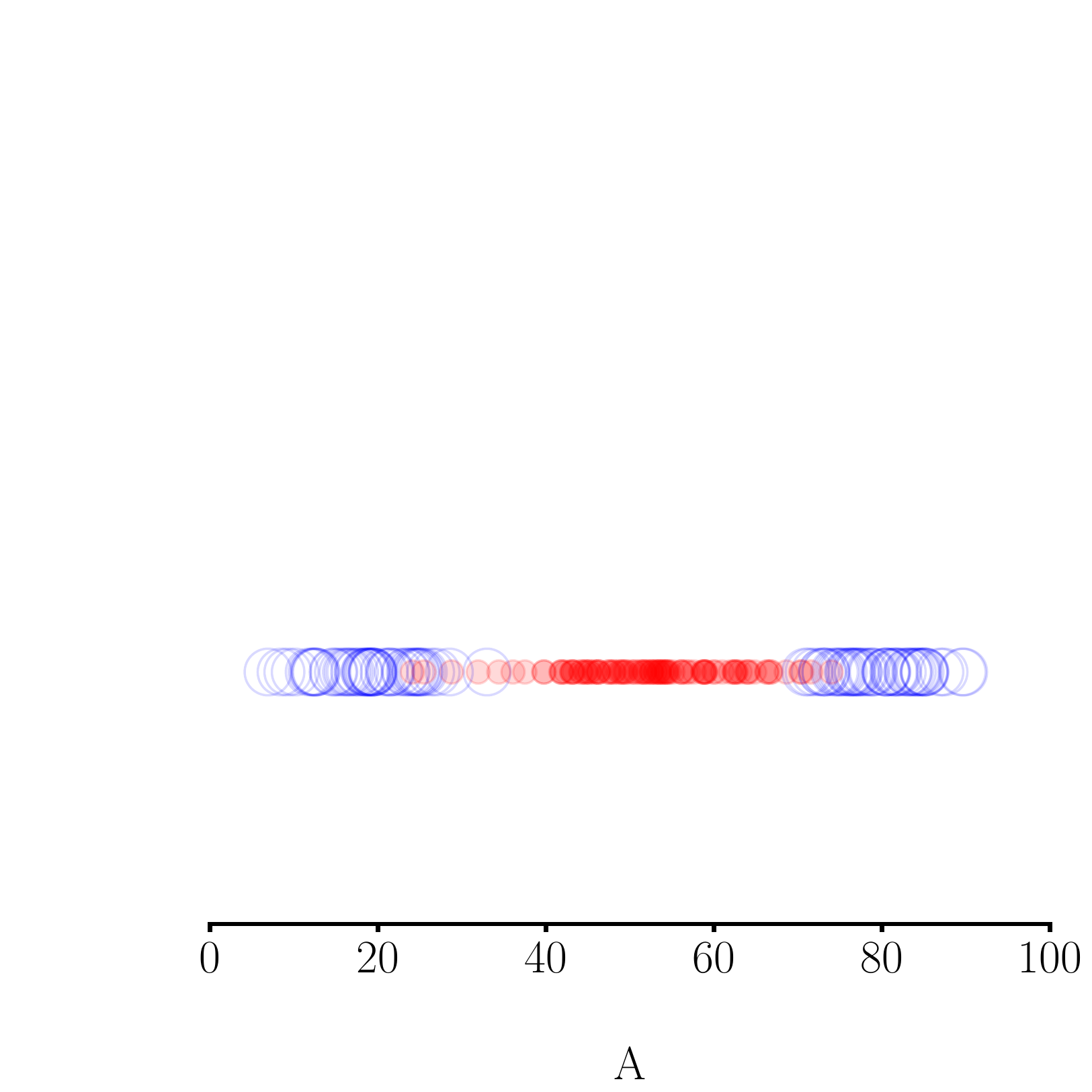}\\
        \includegraphics[scale=\scale,scale=\myscale,trim={0 -1cm -0.5cm 0},clip]{\rootpath 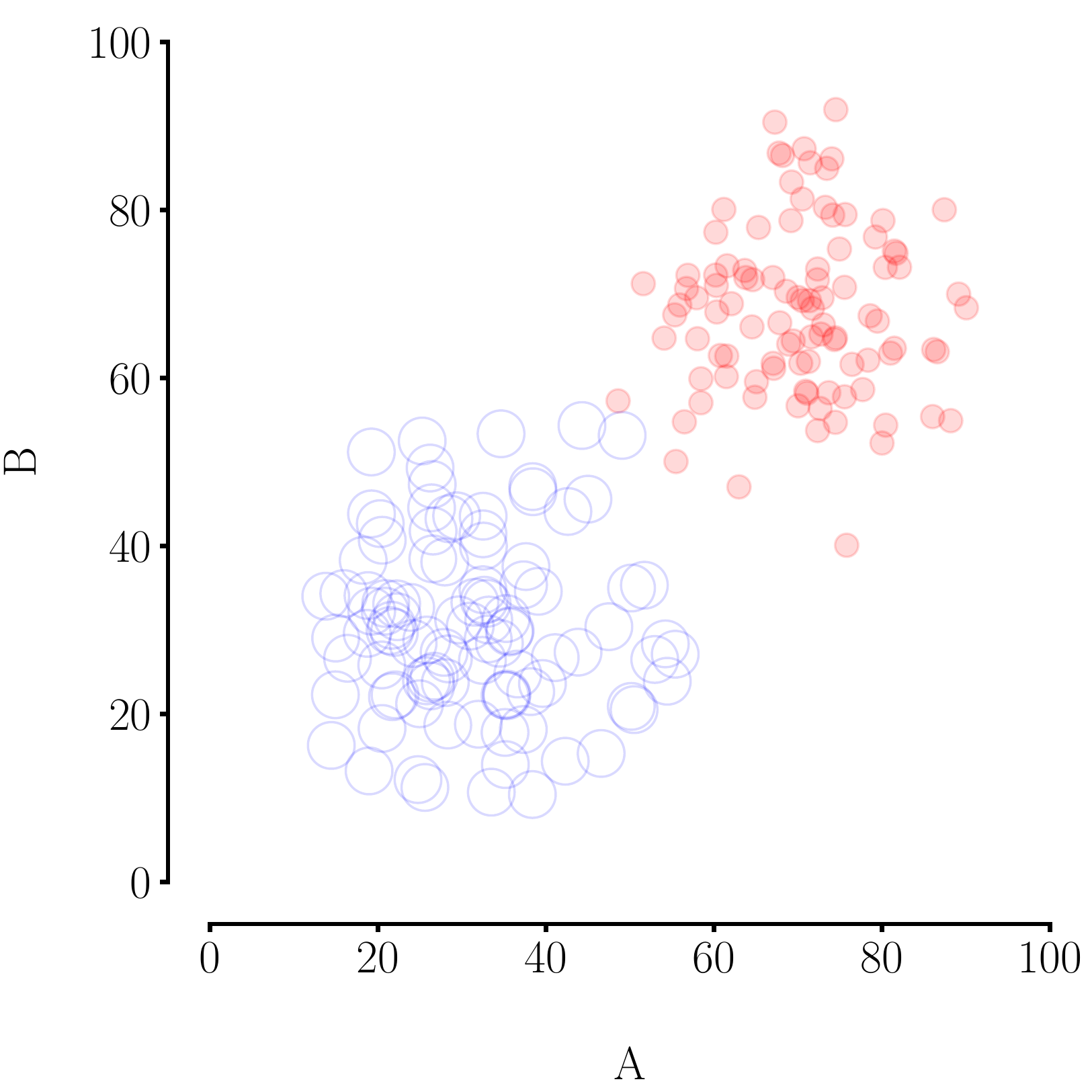}&
        \includegraphics[scale=\scale,scale=\myscale,trim={-0.5cm -1cm 0 0},clip]{\rootpath 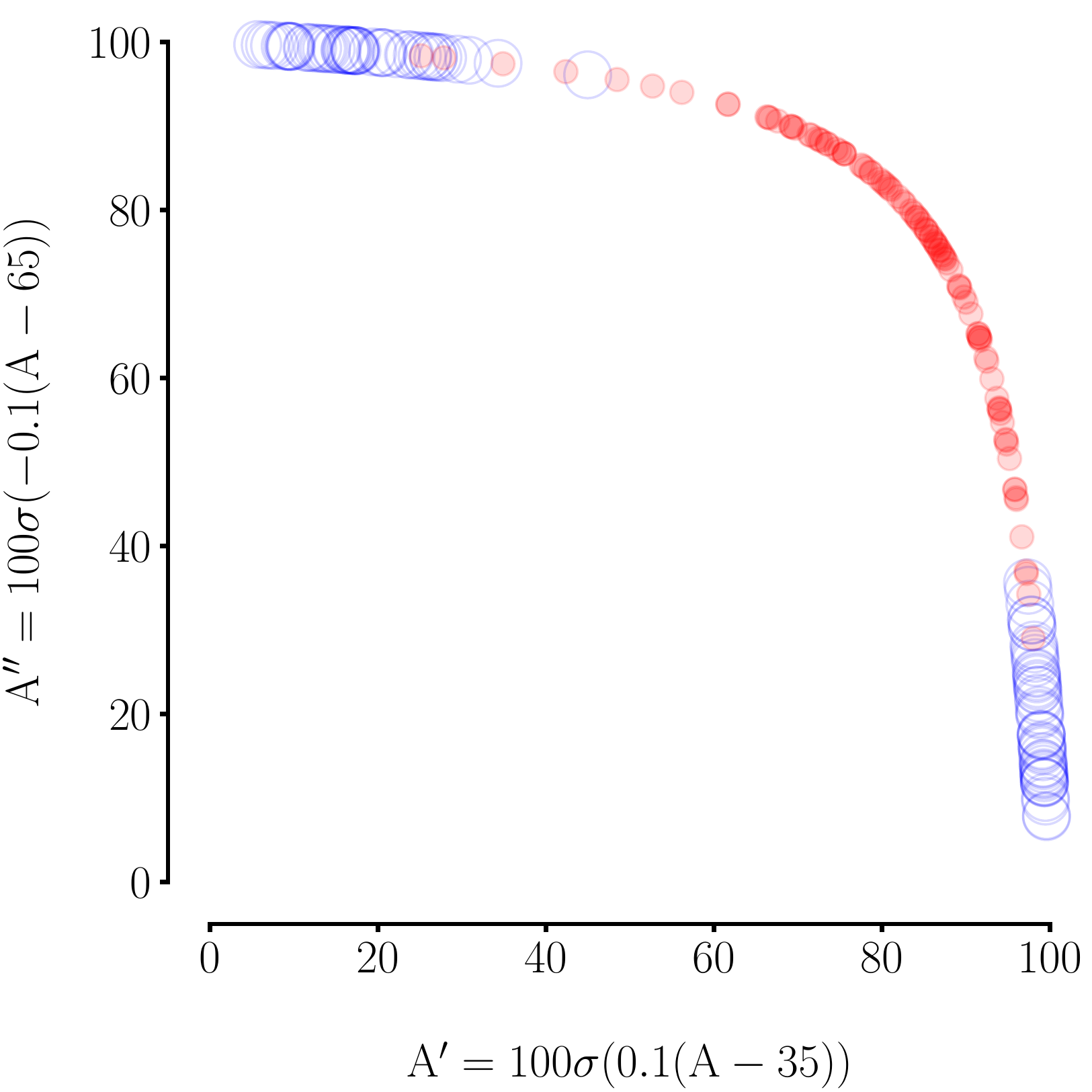}\\
        \includegraphics[scale=\scale,scale=\myscale,trim={0 0 -0.5cm 6cm},clip]{\rootpath 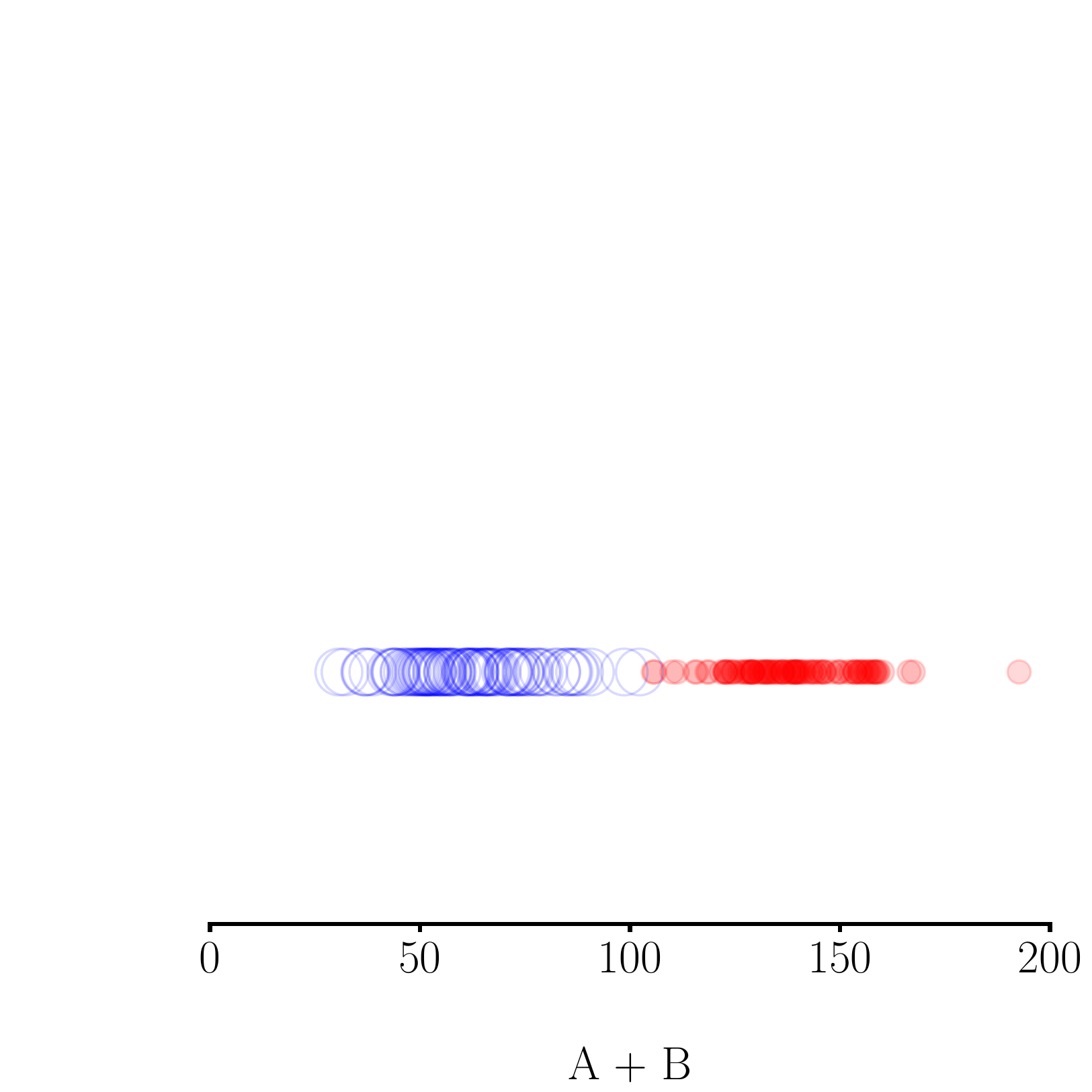}&
        \includegraphics[scale=\scale,scale=\myscale,trim={-0.5cm 0 0 6cm},clip]{\rootpath 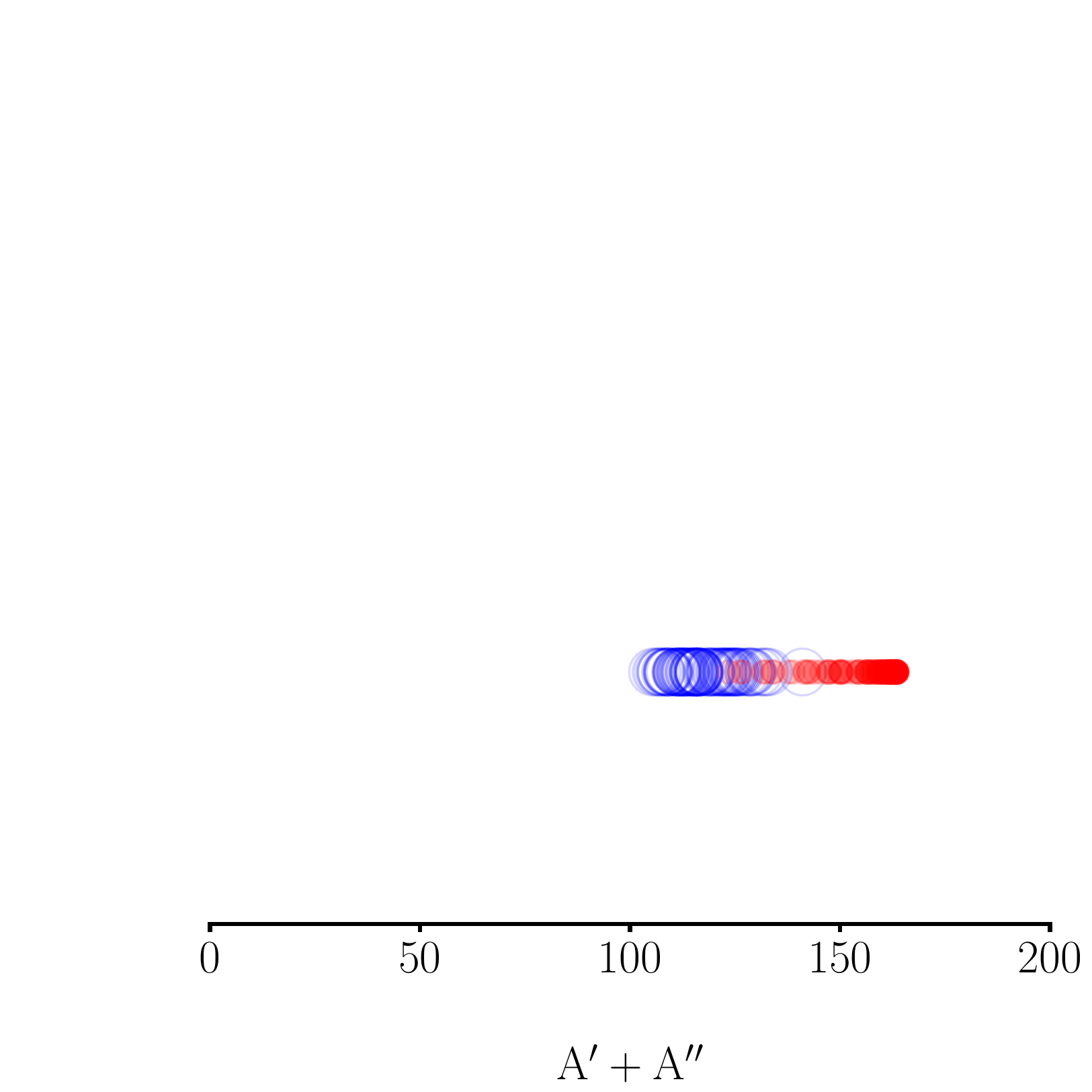}
    \end{tabular}
    \caption[Effect of width and depth of the \gls{cnn}.]{LEFT: Effect of the width of the \gls{cnn}. Two classes (BLUE and RED) may be difficult to separate if only a single feature (Feature A) is available (TOP). However, acquiring another such feature (Feature B) may help to separate the classes in the higher dimensional space (MIDDLE). Even a simple linear combination could then suffice to make the classes separable by a threshold (BOTTOM). RIGHT: Effect of the depth of the \gls{cnn}. Two classes (BLUE and RED) may have a non-linear decision boundary looking at a certain feature (Feature A, TOP). Multiple but different non-linear functions (here $\sigma$ denotes the sigmoid function) may result in linearly separable features (here A' and B') in a higher dimensional space (MIDDLE). Again, a simple linear combination could then suffice to make the classes separable by a threshold (BOTTOM).}
    \label{fig:effect_of_width_and_depth}
\end{figure}

Another educational, though arbitrary, understanding of $m$ is by grouping the first E layers into a feature extraction function $m_e$ and the final C layers into a classification function $m_c$:
\begin{equation}
    m=\underbrace{m^\mathrm{L}\circ m^{\mathrm{L}-1}\circ ...\circ m^{\mathrm{L}-\mathrm{C}+1}}_{m_c}\circ\underbrace{m^\mathrm{E}\circ ...\circ m^2\circ m^1}_{m_e}.
    \label{eq:mls}
\end{equation}
Thus with $m(x)=m_c(m_e(x))=y$. Typical for any layer in $m_c$ is that $\Phi^l(i)=\{i\}$, thus $|\Phi^l(i)|=1$ for $\mathrm{L}-\mathrm{C}+1 \leq l \leq \mathrm{L}$, which means that in the \gls{3d} case only a 1~x~1~x~1 voxel neighborhood is taken into account when calculating the successive feature maps. To understand the action of $m_c$, take $\mathrm{E}=0$ such that our input $x$ is the feature map $y^\mathrm{E}$ that is used by $m_c$ to predict the manual segmentation of the tumor. Further assume that we do not start from the \gls{ct}, but rather from a hypothetical imaging modality in which the measured signal is proportional to how likely that voxel will be part of the tumor. Despite that the values for healthy and tumor tissue might still overlap, looking at Figure~\ref{fig:effect_of_width_and_depth} (LEFT), a single-layer $m_c$ could do the trick using the sigmoid function for $\sigma$ due to the linear separation. Note that the acquisition of another imperfect imaging modality could be used to make the separation more discriminative. On the other hand, if the separation is not linear as in Figure~\ref{fig:effect_of_width_and_depth} (RIGHT), having a preceding layer that computes different linear projections followed by a non-linear activation, might also do the trick. It turns out that this successive application of (multiple) linear projections followed by non-linear activation is at the core of the power of a \gls{cnn}. It intuitively explains how \textit{wider} (i.e., the more features in each layer) and \textit{deeper} (i.e., the more layers) \gls{cnn}s are able to handle more complex tasks. Further note that this process applied the same ``classification function'' at each spatial index $i$ such that segmentation boiled down to \textit{voxel-wise classification}.
\subsection{Feature extraction and the receptive field}
In many cases, it is not realistic to assume that the features produced by the imaging modality, assuming $x=y^\mathrm{E}$ and which typically reflect at each spatial index characteristics that are only influenced by the tissue located at that spatial index, will be discriminative as long as we have a wide and deep classification function $m_c$. To understand the action of $m_e$, we can make the same reasoning for $m_e$ as we did for $m_c$, but instead of acquiring and combining multiple features at a single spatial index $i$, we could combine them from a larger spatial neighborhood around $i$, thus with $|\Phi^l(i)|>1$ for $1 \leq l \leq \mathrm{E}$, to take into account the spatial distribution of features. Indeed, in many segmentation tasks, there is a spatial structure in the data. For example, since a lesion is a connected structure, we could form a more discriminative feature space by including at the spatial index $i$ also the intensities from its surroundings. In the \gls{3d} case, often $|\Phi^l(i)|=27$ in the form of a 3~x~3~x~3 voxel neighborhood is used. As a result, if E increases, the patterns that are used in $m_c$ come from a more global region in $x$. This is what is called the \textit{receptive field} (Figure~\ref{fig:receptive_field}). We want to stress that the separation of $m$ into $m_e$ and $m_c$ is arbitrary and ambiguous. In practice, often a layer in $m_e$ is called a convolutional layer and a layer in $m_c$ is called a fully-connected or dense layer. However, they do not need to be separated into different types, since both perform convolutions and layers in $m_c$ are in fact less ``connected'' than the layers in $m_e$. At the core of the \gls{cnn} is that the same patterns are looked for across the entire image and that the complexity and receptive field of the patterns can grow simultaneously. With respect to the receptive field, there are two more things to note. First, it is possible that the receptive field in a certain layer extends beyond the spatial indices present in $x^l$. In practice, either such a scenario is avoided by having a larger $x$ than $y$, or by padding, usually with zeros, such that the sizes of $x$ and $y$ remain identical. Second, to obtain a more aggressive increase of the receptive field, often so-called pooling or strided convolutions are used. In both cases, the original assumption that the spatial indices $i \in \mathcal{I}$ in $y$ are present in $x$ remains valid for all layers. When looking at Figure~\ref{fig:receptive_field}, imagine how the implementation of pooling and upconvolution (e.g., via sparse or strided convolutions) do effect this representation.
\section{Sampling the set of data pairs}
Having introduced the action of a basic \gls{cnn} for segmentation, it is time to discuss the set of data pairs.
It was mentioned before that the \gls{cnn} minimizes the loss over a given set of examples. However, what is this set of examples and why is it important to keep track of? In essence, we can view the input $X$ and output $Y$ as random variables, and thus with $x$ and $y$ denoting realizations thereof. During training, the \gls{cnn} has seen a finite set of N realizations in $\mathcal{S}_\mathrm{train}=\{(x_n,y_n)\}_{1 \leq n \leq \mathrm{N}}$ from which the empirical distribution $P_\mathrm{train}(X,Y)$ arises. Since the \gls{cnn} can be trained to transform $X$ into $Y$, it can be good practice to monitor what $X$ and $Y$ represent, such that during testing we can aim for an $\mathcal{S}_\mathrm{test}$ with $P_\mathrm{test}(X,Y) \approx P_\mathrm{train}(X,Y)$ (or vice versa).
\subsection{A composite sampling scheme}
Before, we had assumed, somewhat naively, that the input $X$ was the \gls{ct} and the output $Y$ was the manual segmentation of the tumor. However, this is mostly not true from the learning perspective. Let us start from the entire space of possible \gls{ct} images with accompanying manual tumor segmentations, say $P(X_\mathrm{\gls{ct}},Y_\mathrm{seg})$. This joint distribution embodies all sorts of variability, e.g., due to different patient, acquisition, and expert characteristics, but also preprocessing may be included. The first sampling step $s'$ results in a set of images from subjects and is inevitably biased, e.g., towards a specific country or data from a certain clinical trial. The second sampling step $s''$ is due to practical limitations such as the available computational memory and the receptive field property of the \gls{cnn}. The former may result in the sampling of so-called \textit{patches} of a size smaller than the original image. The latter, depending on the padding strategy, further modifies how $X$ will look like. Hence, we arrive at the following composite sampling scheme:
\begin{equation}
    \mathcal{S}=\{(x_n,y_n)\}_{1 \leq n \leq \mathrm{N}} \xleftarrow[]{s''} \{(x_{\mathrm{ct},m},y_{\mathrm{seg},m})\}_{1 \leq m \leq \mathrm{M}} \xleftarrow[]{s'} P(X_\mathrm{\gls{ct}},Y_\mathrm{seg}).
    \label{eq:sampling}
\end{equation}
As mentioned before, it is important to keep track of this sampling process to make the \gls{cnn} generalize well from training to testing fases.
\subsection{Further note on spatial origin}
\begin{figure}[t!]
    \newcommand\myscale{\scaleppt} 
    \setlength\tabcolsep{0pt}
    \centering
    \begin{tabular}{ll}
        \includegraphics[scale=\scale,scale=\myscale,trim={0 -1cm -1cm 0},clip]{\rootpath 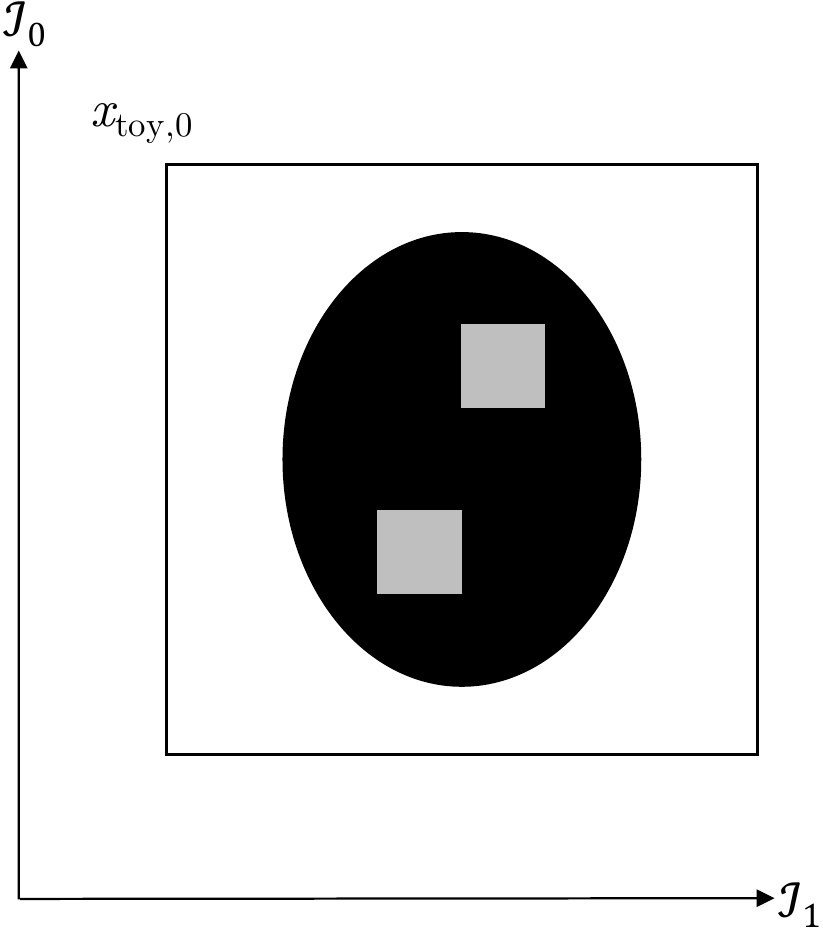}&
        \includegraphics[scale=\scale,scale=\myscale,trim={0 -1cm 0 0},clip]{\rootpath 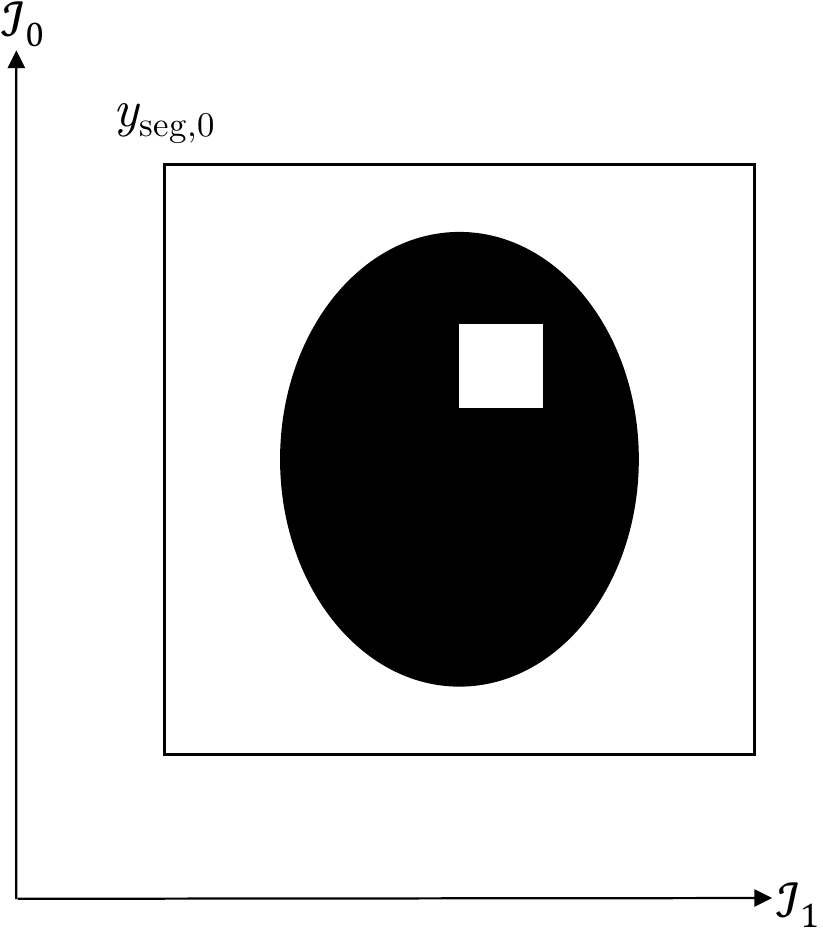}\\
        \includegraphics[scale=\scale,scale=\myscale,trim={0 -1cm -1cm 0},clip]{\rootpath 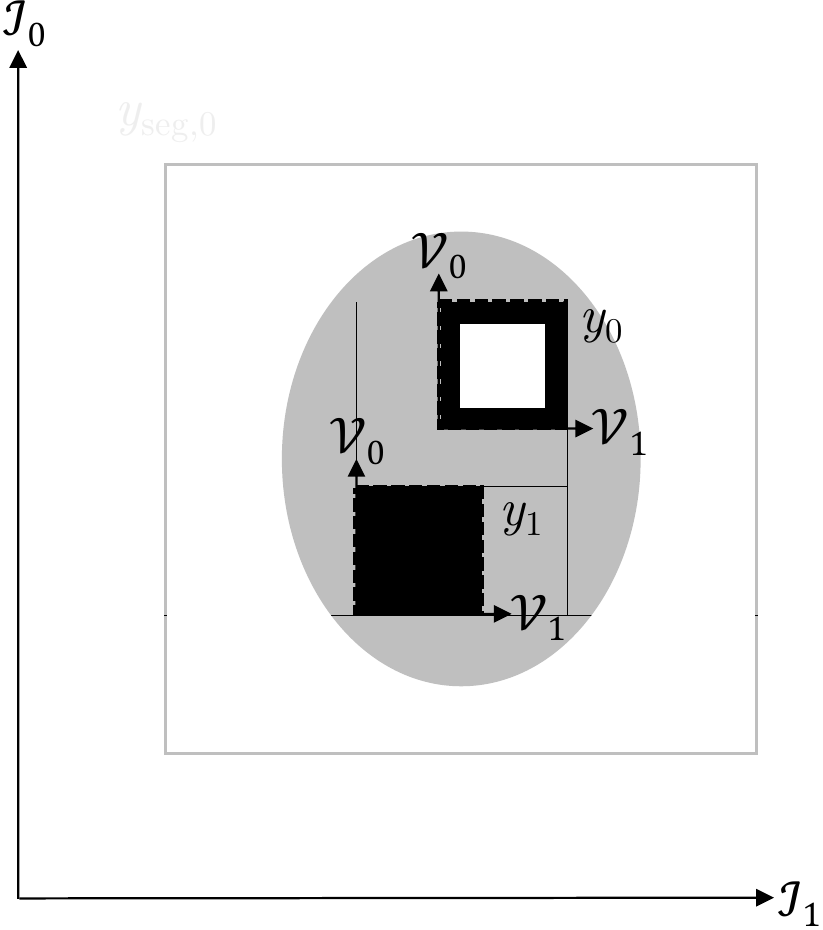}&
        \includegraphics[scale=\scale,scale=\myscale,trim={0 -1cm 0 0},clip]{\rootpath 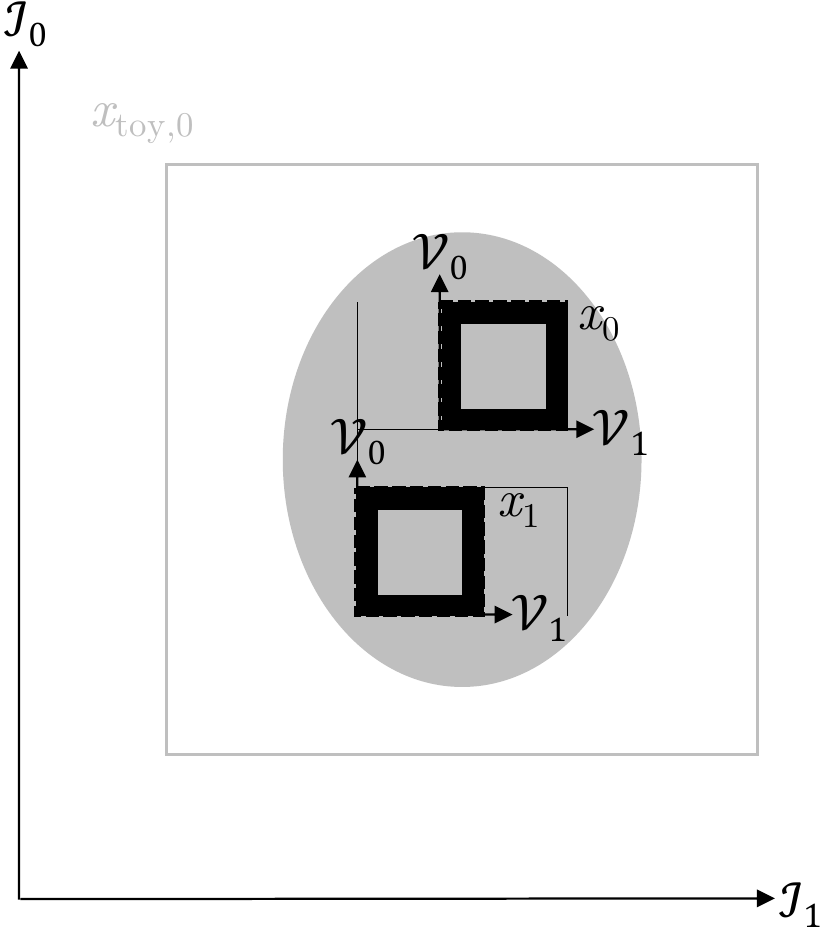}\\
        \includegraphics[scale=\scale,scale=\myscale,trim={0 0 -1cm 0},clip]{\rootpath 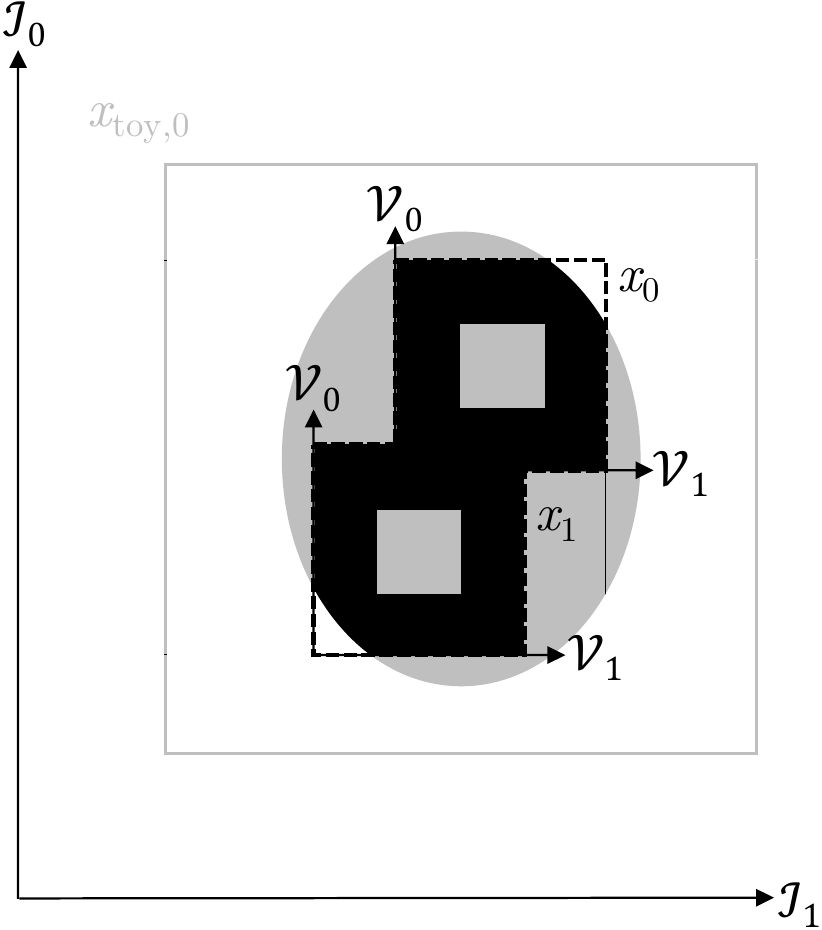}&
        \includegraphics[scale=\scale,scale=\myscale]{\rootpath 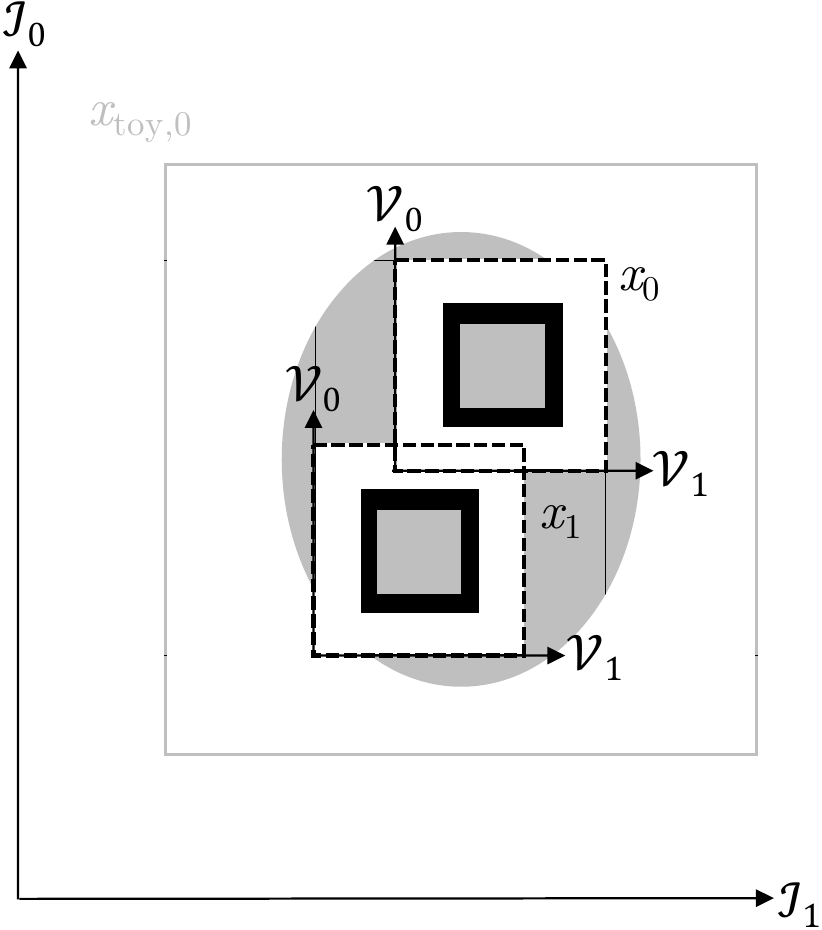}
    \end{tabular}
    \caption[Importance of the spatial origin of the data.]{TOP: imagine a toy segmentation problem in which we only want to segment the top-right lesion. MIDDLE-LEFT: suppose our learning set involves the learning form image patches. MIDDLE-RIGHT: with a \gls{rf}~=~1~x~1 we can see that it lost the information needed. BOTTOM-LEFT: with a larger \gls{rf} and padding with image data, part of the patch will be possible to descriminate. BOTTOM-RIGHT: even with a larger \gls{rf} the information might be insufficient due to zero padding.}
    \label{fig:patch_sizes}
\end{figure}
It is especially the second sampling step $s''$ that deserves some extra attention. The spatial axis $\mathcal{I}$ can be thought of as the set of ``locations in the world''. By converting the original data to $\mathcal{S}$, we convert the data to a format that can be used by the \gls{cnn}, but we must be aware of two things. First, the interplay between patch size, receptive field size and padding may result in information loss such that the task cannot be performed anymore. This is explained in more detail in Figure~\ref{fig:patch_sizes}. Second, we must remember where the patch came from in our original image, thus in $\mathcal{I}$, in order to be able to put back the patch with its own relative voxel axis $\mathcal{V}$ correctly. These transformations are mostly affine, but due to successive transformations the overall transformation may become lost. To overcome this burder we developed the DeepVoxNet2 framework~\cite{deepvoxnet2} in which the overall transformation is tracked.
\section{State-of-the-art CNN architectures}
We need to make the historical connection with image classification to connect with recent \gls{cnn} architectures for segmentation. Many breakthroughs were made in image classification and were only later adopted in segmentation. This is possible because a \gls{cnn} for voxel-wise classification can be used straightforwardly for image classification and vice versa. In image classification, we have only a single label per image. Hence, padding is less of an issue and we can view image classification as segmenting only the central voxel in the output (you can imagine this in Figure~\ref{fig:receptive_field}). The latter imposes a soft requirement on the \gls{cnn} architecture to look for patterns across a large part of $x$.
\subsection{Historical work in image classification}
The introduction of neural networks can be traced back to~\cite{Fukushima1980} and was inspired by the hierarchical pattern recognition in the visual cortex. Soon, backpropagation of gradients was introduced as an effective way to learn the internal weights of a \gls{cnn}~\cite{Lecun1998}. However, it took some years before the computational power, computer memory, and available data matched the intrinsic capacity of \gls{cnn}s. It was not until 2012, with the advent of graphics processing unit computing and the use of a large-scale dataset (i.e., ImageNet 2012~\cite{Russakovsky2015a}) that the so-called AlexNet~\cite{Krizhevsky2012} was able to put \gls{cnn}-based implementations on the map. Using AlexNet, we can quickly identify some crucial aspects and connect with specific alterations in more recent \gls{cnn}s. It turns out that the AlexNet was not only the first good architecture (e.g., depth, width, \gls{rf}) but that it also was a clever integration of multiple disperse improvements that are now still at the core of state-of-the-art \gls{cnn}s.\\
First, with $\mathrm{E}=8$ (including 3 max pooling layers~\cite{Goodfellow2013}) and $\mathrm{C}=3$ the \gls{cnn} was architectured to have a \gls{rf} of 254~x~254 and as such was able to extract features from across a large part of a 256~x~256 image. While its width and depth were key ingredients, it did not take long for some to change. In 2014, with the VGG-nets, it was shown that \gls{cnn} architectures benefit from having smaller kernel sizes (i.e., 3~x~3) and more layers~\cite{Simonyan2014}. Intuitively, smaller filters in each layer result in fewer parameters, such that wider and deeper networks, and thus more complexity, can be obtained. Furthermore, using the complexity and scale of each feature map, only local patterns are useful and generalizable across the spatial axis.\\
Second, since the \gls{rf} of AlexNet was smaller than the image size, they could augment $S$ by sampling nine spatial indices in $s''$ and, as such, implement translation as a straightforward technique for data augmentation. Furthermore, the function $s''$ included two other label-preserving transformations to further extent the variety in $S$ to prevent overfitting: horizontal flipping and color intensity alteration (here based on principle component analysis). Soon, even more aggressive data augmentation techniques were used, including affine and elastic distortions, which are still used today to make $S$ even more diverse.\\
Third, the \gls{relu} function was used as the non-linearity $\sigma$ (instead of, e.g., the sigmoid function) in the intermediate layers~\cite{Nair2010} to combat vanishing gradients and is now still used in the form of leaky or parametric ReLU functions~\cite{Maas2013}. Similarly, in 2015, the so-called ResNets~\cite{He2015} introduced skip connection to aid the learning process by letting gradients bypass layers during backpropagation.\\
Fourth, the feature maps $x^l=y^{l-1}$ in the intermediate layers were normalized using what they called local response normalization. Roughly, we can link to current implementations using the attention mechanism~\cite{Wang2017} (in this case channel-wise) and group normalization~\cite{Wu2018} (in this case calculated at every spatial index; in the limit also known as layer normalization~\cite{Ba2016}). Other frequent normalizations are batch normalization~\cite{Ioffe2015} and instance normalization~\cite{Ulyanov2016}, depending on which axes in $x^l$ are used to calculate the normalization parameters.\\
Fifth, AlexNet also uses dropout~\cite{Hinton2012} in the first two layers in $m_c$ as an alternative way to reduce overfitting by preventing co-adaptation of neurons. Additionally, they also used L2-norm regularization on the trainable weights, which in combination with L1-norm regularization is still common in recent implementations. L2-norm regularization seems to be particularly useful in preventing a virtual decrease in the learning rate when using one of the normalizations mentioned above ~\cite {Li2019a}.\\
Finally, in AlexNet, the optimizer (i.e., how a gradient is turned into a weight update) was simple stochastic gradient descent with momentum. Currently, more recent optimizers, such as ADAM~\cite{Kingma2014a} and RMSProp~\cite{Tielemans2012}, implement an adaptive learning rate for each trainable weight based on its current magnitude.
\subsection{Historical work in image segmentation}\label{sec:history_segmentation}
\begin{figure}[t!]
    \newcommand\myscale{\scaleppt} 
    \setlength\tabcolsep{0pt}
    \centering
    \begin{tabular}{c}
        \includegraphics[scale=\scale,scale=\myscale,trim={0 -1cm 0 0},clip]{\rootpath 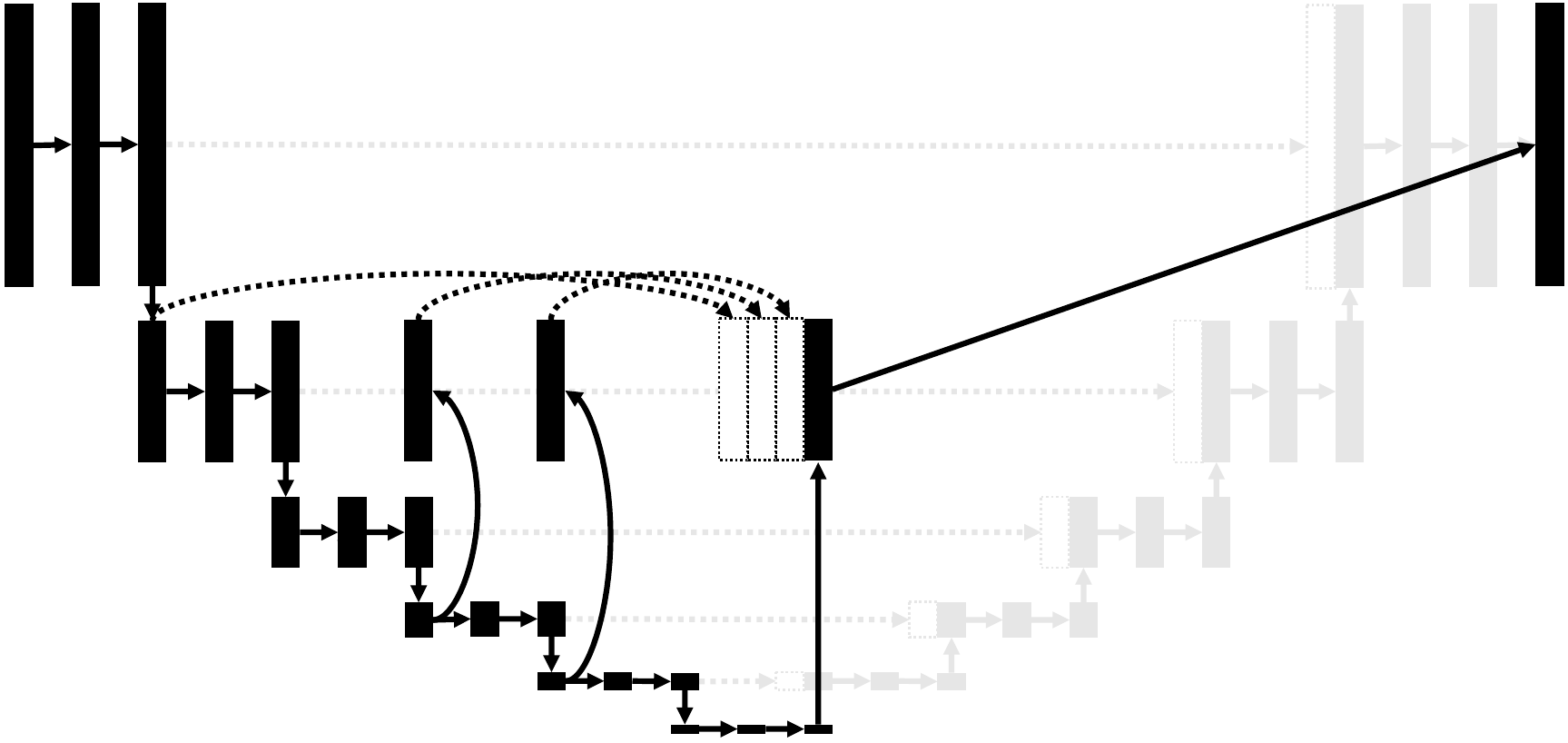}\\
        \includegraphics[scale=\scale,scale=\myscale,trim={0 -1cm 0 0},clip]{\rootpath 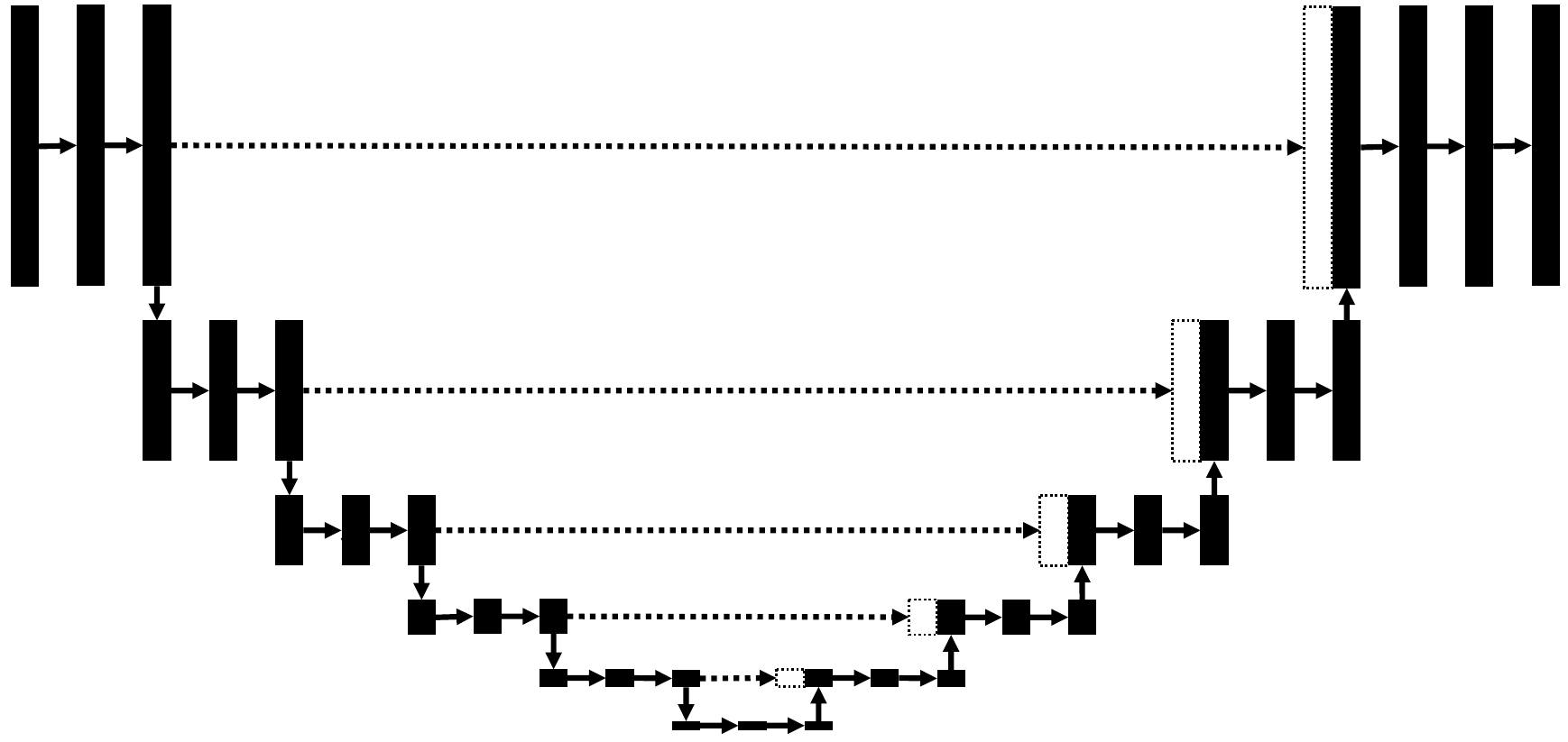}\\
        \includegraphics[scale=\scale,scale=\myscale]{\rootpath 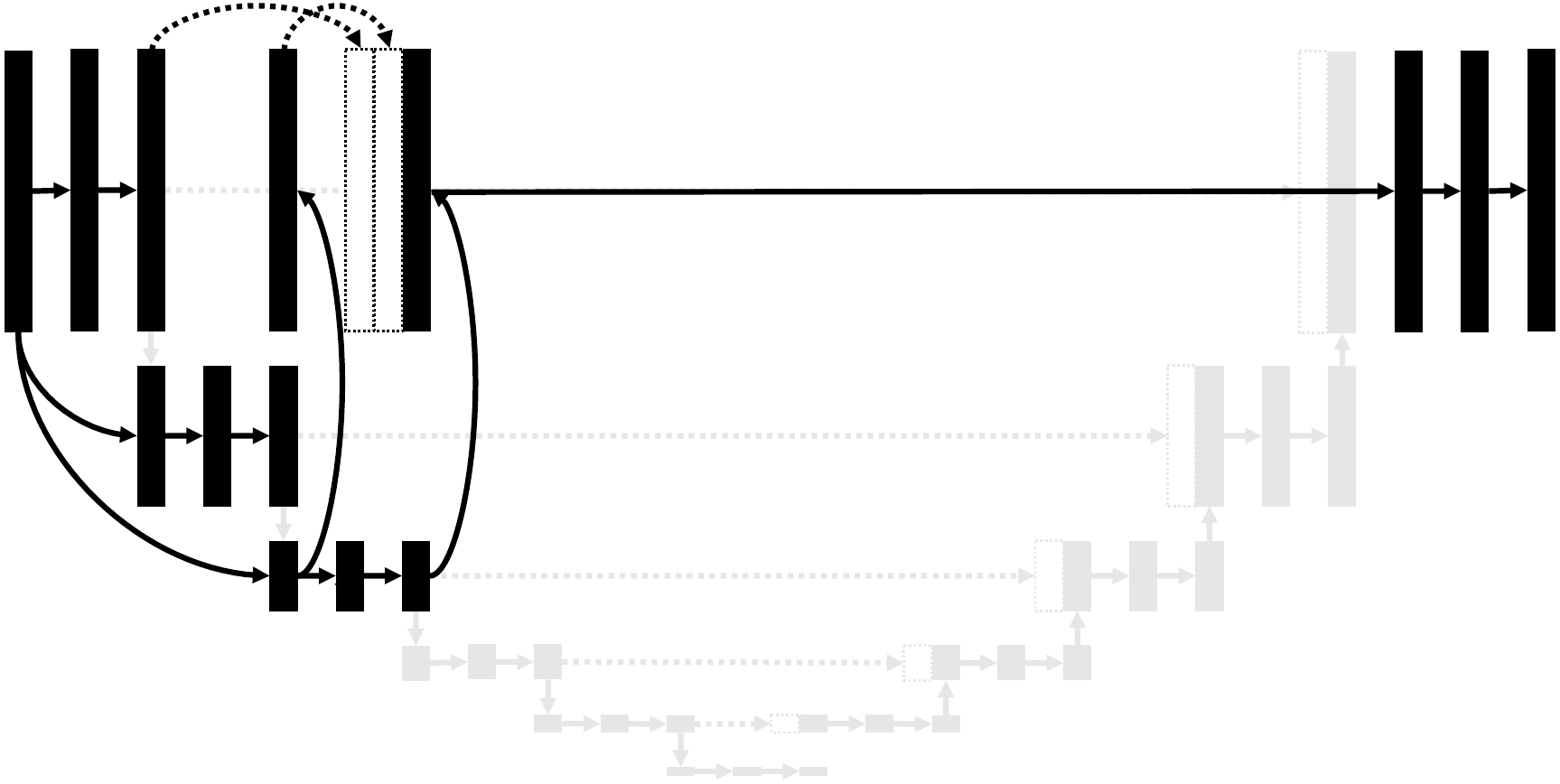}
    \end{tabular}
    \caption[Schematic comparison of FCN, U-Net and DeepMedic.]{Schematic comparison of a FCN-like (TOP, here it is FCN-2s), U-Net-like (MIDDLE) and DeepMedic-like (BOTTOM) architecture. Note that ``-like'' refers to the fact that not necessarily the same number of features, layers, pooling/upsampling steps are used.}
    \label{fig:architecture_comparison}
\end{figure}
The first \gls{cnn}-based segmentation method used the voxel-wise classification explicitly~\cite{Ciresan2012}. As a result, its novelty is somewhat limited in the sense that it uses a similar architecture as in AlexNet but $\mathcal{S}$ is different and trivially expanded to sample all voxels using $s''$.\\
In 2014, a more efficient implementation was introduced in which the effects of pooling were obtained using sparse convolutions~\cite{Li2014}. It became possible to have an output patch size larger than 1, without any difference in the voxel-wise learning paradigm. Intuitively, a feature in a particular feature map only depends on the spatial index $i$ in $\mathcal{I}$ and not where that would be in the relative axis $\mathcal{V}$. One drawback of this implementation was the increased memory requirement. If we take a fixed architecture and modify the operations to sparse convolutions and pooling, the size of every intermediate $y^l$ will increase. If the explicit voxel-wise implementation took up all memory, we could not benefit from this computational improvement. In fact, we could only harvest the computational benefits by making the \gls{cnn} smaller (e.g., reducing the depth, width or $\mathrm{\gls{rf}}$), and this was in sharp contrast with the parallel observations in image classification, including the  rise of the deeper VGG-nets mentioned above.\\
In 2015, the so-called fully-convolutional networks (\gls{fcn}s)~\cite{Long2015} came to the rescue. Instead of using sparse convolutions, they kept the original implementation of pooling in combination with a new and learnable upsampling operation, i.e., transposed convolution. The simplest variant (i.e. the \gls{fcn}-32s) learned a 32-times upsampling right from the final prediction. This was further improved by a cascaded upsampling by first upsampling the most coarse prediction to and combining it with the prediction on an intermediate resolution, and only then upsampling to the original image resolution (i.e., the \gls{fcn}-16s and \gls{fcn}-8s). As a result, similar to the advancement above of residual connections in image classification, skip connections were introduced as a secondary effect in segmentation. Note that the final layer now performs a convolution while taking up the role of the $m_c$ with $\mathrm{C}=1$ (these are now called dense layers).\\
From the \gls{fcn} it was only a small step to the king and queen in medical imaging: U-Net~\cite{Ronneberger2015} and DeepMedic~\cite{Kamnitsas2017} (Figure~\ref{fig:architecture_comparison}). In both U-Net and DeepMedic, instead of upsampling (intermediate) predictions, now (intermediate) feature maps were upsampled directly and followed by several dense layers with $\mathrm{C}>1$. In U-Net, the cascaded upsampling of a virtual \gls{fcn}-1s had become more gradual, and thus the decoder had become deeper. In DeepMedic, the decoder was kept unchanged (although the transposed convolution was replaced by 0-th order upsampling), but the encoder was modified such that features on a coarser scale were calculated starting from the input image at a lower resolution. As such, they could further reduce the sizes of the intermediate $y^l$ and reduce the trade-off between network depth and output patch size of $y^L$.\\
The \gls{fcn}, U-Net and DeepMedic, contribute their improvements over the original voxel-wise implementations to a superior memory-complexity trade-off and skip connections. However, we believe another aspect of being the true motor behind their popularity: the implicit data augmentation of the features calculated for each spatial index. Imagine we have a single pooling layer, and thus that we look for features on two scales. Features on the original scale are calculated for each spatial index in $x$, and, therefore, can be considered centered at each spatial index. However, features on the coarser scale are calculated only for a sparse number of spatial indices with respect to the first scale. As a result, those features are only exact for a subset of spatial indices in the original scale. After upsampling, these features are interpolated. As a result, the architecture hard codes features on a coarser scale to be invariant with respect to a finer scale. And there is another implicit data augmentation going on as soon as the spatial size of any $y^l$ increases. Similar to image classification, batch, instance, and group normalization have found their way to segmentation. The parameters for all three normalization schemes will depend on the spatial indices present in $y^l$. As a result, the features used to classify a spatial index $i$ will vary across different $y_n$ that contain $i$.
\section{Conclusion}
In this article, we looked into some essential aspects of convolutional neural networks (\gls{cnn}s) with the focus on medical image segmentation. First, we discussed the \gls{cnn} architecture, thereby highlighting the spatial origin of the data, voxel-wise classification and the receptive field. Second, we discussed the sampling of input-output pairs, thereby highlighting the interaction between voxel-wise classification, patch size and the receptive field. Finally, we gave a historical overview of crucial changes to \gls{cnn} architectures for classification and segmentation, giving insights in the relation between three pivotal \gls{cnn} architectures: FCN~\cite{Long2015}, U-Net~\cite{Ronneberger2015} and DeepMedic~\cite{Kamnitsas2017}.
\bibliographystyle{plain}
\bibliography{references}

\begin{thebibliography}{10}

\bibitem{Ba2016}
Jimmy~Lei Ba, Jamie~Ryan Kiros, and Geoffrey~E. Hinton.
\newblock {Layer Normalization}.
\newblock {\em arXiv}, July 2016.

\bibitem{Ciresan2012}
Dan~C. Cireşan, Alessandro Giusti, Luca~M. Gambardella, and J{\"{u}}rgen
  Schmidhuber.
\newblock {Deep Neural Networks Segment Neuronal Membranes in Electron
  Microscopy Images}.
\newblock In {\em Advances in Neural Information Processing Systems}, volume~4,
  pages 2843--2851. Curran Associates, Inc., 2012.

\bibitem{deepvoxnet2}
DeepVoxNet2.
\newblock {https://github.com/JeroenBertels/deepvoxnet2}.

\bibitem{Fukushima1980}
Kunihiko Fukushima.
\newblock {Neocognitron: A self-organizing neural network model for a mechanism
  of pattern recognition unaffected by shift in position}.
\newblock {\em Biological Cybernetics}, 36:193--202, 1980.

\bibitem{Goodfellow2013}
Ian~J. Goodfellow, David Warde-Farley, Mehdi Mirza, Aaron Courville, and Yoshua
  Bengio.
\newblock {Maxout Networks}.
\newblock {\em arXiv}, pages 1319--1327, 2013.

\bibitem{He2015}
Kaiming He, Xiangyu Zhang, Shaoqing Ren, and Jian Sun.
\newblock {Delving Deep into Rectifiers: Surpassing Human-Level Performance on
  ImageNet Classification}.
\newblock {\em arXiv}, pages 1--11, 2015.

\bibitem{Hinton2012}
Geoffrey~E. Hinton, Nitish Srivastava, Alex Krizhevsky, Ilya Sutskever, and
  Ruslan~R. Salakhutdinov.
\newblock {Improving neural networks by preventing co-adaptation of feature
  detectors}.
\newblock {\em arXiv}, pages 1--18, July 2012.

\bibitem{Ioffe2015}
Sergey Ioffe and Christian Szegedy.
\newblock {Batch Normalization: Accelerating Deep Network Training by Reducing
  Internal Covariate Shift}.
\newblock {\em arXiv}, pages 1--11, February 2015.

\bibitem{Kamnitsas2017}
Konstantinos Kamnitsas, Christian Ledig, Virginia~F.J. Newcombe, ..., David~K.
  Menon, Daniel Rueckert, and Ben Glocker.
\newblock {Efficient multi-scale 3D CNN with fully connected CRF for accurate
  brain lesion segmentation}.
\newblock {\em Medical Image Analysis}, 36:61--78, February 2017.

\bibitem{Kingma2014a}
Diederik~P. Kingma and Jimmy Ba.
\newblock {Adam: A Method for Stochastic Optimization}.
\newblock {\em arXiv}, pages 1--15, December 2014.

\bibitem{Krizhevsky2012}
Alex Krizhevsky, Ilya Sutskever, and Geoffrey~E Hinton.
\newblock {ImageNet classification with deep convolutional neural networks}.
\newblock {\em Communications of the ACM}, 60(6):84--90, May 2017.

\bibitem{Lecun1998}
Yann LeCun, L.~Bottou, Y.~Bengio, and P.~Haffner.
\newblock {Gradient-Based Learning Applied to Document Recognition}.
\newblock {\em Proceedings of the IEEE}, 86(11):2278 -- 2324, 1998.

\bibitem{Li2014}
Hongsheng Li, Rui Zhao, and Xiaogang Wang.
\newblock {Highly Efficient Forward and Backward Propagation of Convolutional
  Neural Networks for Pixelwise Classification}.
\newblock {\em arXiv}, 2014.

\bibitem{Li2019a}
Zhiyuan Li and Sanjeev Arora.
\newblock {An Exponential Learning Rate Schedule for Deep Learning}.
\newblock {\em arXiv}, pages 1--29, October 2019.

\bibitem{Long2015}
Jonathan Long, Evan Shelhamer, and Trevor Darrell.
\newblock {Fully convolutional networks for semantic segmentation}.
\newblock In {\em IEEE Conference on Computer Vision and Pattern Recognition
  (CVPR)}, pages 3431--3440. IEEE, June 2015.

\bibitem{Maas2013}
Andrew~L Maas, Awni~Y Hannun, and Andrew~Y Ng.
\newblock {Rectifier nonlinearities improve neural network acoustic models}.
\newblock {\em in ICML Workshop on Deep Learning for Audio, Speech and Language
  Processing}, 28, 2013.

\bibitem{Nair2010}
Vinod Nair and Geoffrey~E Hinton.
\newblock {Rectified Linear Units Improve Restricted Boltzmann Machines}.
\newblock {\em International Conference on Machine Learning}, 27(3):807--814,
  2010.

\bibitem{Ronneberger2015}
Olaf Ronneberger, Philipp Fischer, and Thomas Brox.
\newblock {U-Net: Convolutional Networks for Biomedical Image Segmentation}.
\newblock In {\em MICCAI}, pages 234--241. Springer Nature, 2015.

\bibitem{Russakovsky2015a}
Olga Russakovsky, Jia Deng, Hao Su, ..., Michael Bernstein, Alexander~C. Berg,
  and Li~Fei-Fei.
\newblock {ImageNet Large Scale Visual Recognition Challenge}.
\newblock {\em International Journal of Computer Vision}, 115:211--252, 2015.

\bibitem{Simonyan2014}
Karen Simonyan and Andrew Zisserman.
\newblock {Very Deep Convolutional Networks for Large-Scale Image Recognition}.
\newblock {\em arXiv}, pages 1--14, September 2014.

\bibitem{Tielemans2012}
T.~Tielemans and Geoffrey~E. Hinton.
\newblock {Lecture 6.5: RMSProp: Divide the gradient by a running average of
  its recent magnitude}.
\newblock In {\em Coursera: Neural Networks for Machine Learning}, 2012.

\bibitem{Ulyanov2016}
Dmitry Ulyanov, Andrea Vedaldi, and Victor Lempitsky.
\newblock {Instance Normalization: The Missing Ingredient for Fast
  Stylization}.
\newblock {\em arXiv}, July 2016.

\bibitem{Wang2017}
Fei Wang, Mengqing Jiang, Chen Qian, ..., Honggang Zhang, Xiaogang Wang, and
  Xiaoou Tang.
\newblock {Residual Attention Network for Image Classification}.
\newblock In {\em IEEE Conference on Computer Vision and Pattern Recognition
  (CVPR)}, pages 6450--6458. IEEE, July 2017.

\bibitem{Wu2018}
Yuxin Wu and Kaiming He.
\newblock {Group Normalization}.
\newblock In {\em Lecture Notes in Computer Science}, volume 11217 LNCS, pages
  3--19. Springer, 2018.

\end{thebibliography}
\end{document}